\documentclass[runningheads]{llncs}
\usepackage{graphicx}
\usepackage{wrapfig}
\usepackage{lipsum}
\usepackage[hidelinks]{hyperref}
\usepackage{xcolor}
\hypersetup{
    colorlinks,
    linkcolor={red!},
    citecolor={green!},
    urlcolor={magenta!}
}

\usepackage{tikz}
\usepackage{comment}
\usepackage{amsmath,amssymb} % define this before the line numbering.
\usepackage{algorithm2e, algorithmic}
\usepackage{color}

\usepackage[accsupp]{axessibility}  % Improves PDF readability for those with disabilities.

\usepackage[width=122mm,left=12mm,paperwidth=146mm,height=193mm,top=12mm,paperheight=217mm]{geometry}

\usepackage{booktabs}

\DeclareMathOperator*{\argmin}{arg\,min}

\begin{document}
% \linenumbers
\pagestyle{headings}
\mainmatter
\def\ECCVSubNumber{16}  % Insert your submission number here

\title{QFT: Post-training quantization via fast joint finetuning of all degrees of freedom} % Replace with your title

%\begin{comment}
% \titlerunning{ECCV-22 submission ID \ECCVSubNumber} 
% \authorrunning{ECCV-22 submission ID \ECCVSubNumber} 
% \author{Anonymous ECCV/CADL2022 submission}
% \institute{Paper ID \ECCVSubNumber}
%\end{comment}
%******************

%\begin{comment}
\titlerunning{QFT: all-DoF Quant-aware Finetuning}
%\titlerunning{Post-training quantization baseline via joint fine-tuning of all degrees-of-freedom}
%\author{Alex Finkelstein\inst{1}\orcidID{0000-1111-2222-3333} \and
\author{Alex Finkelstein\and
Ella Fuchs \and
Idan Tal\and \\
Mark Grobman\and
Niv Vosco\and
Eldad Meller}
\authorrunning{A. Finkelstein et al.}
\institute{{Hailo, Tel-Aviv, Israel} \\ \url{www.hailo.ai} \\ \email{\{alexf,eldadm\}@hailo.ai}}
%\end{comment}
%******************
\maketitle

\begin{abstract}
The post-training quantization (PTQ) challenge of bringing quantized neural net accuracy close to original has drawn much attention driven by industry demand. Many of the methods emphasize optimization of a specific degree-of-freedom (DoF), such as quantization step size, preconditioning factors, bias fixing, often chained to others in multi-step solutions. Here we rethink quantized network parameterization in HW-aware fashion, towards a unified analysis of all quantization DoF, permitting for the first time their joint end-to-end finetuning. Our single-step simple and extendable method, dubbed quantization-aware finetuning (QFT), achieves 4b-weights quantization results on-par with SoTA within PTQ constraints of speed and resource.
\end{abstract}

\section{Introduction} Quantization is a standard step \cite{Hubara17,Jacob17,Krishna,NagelWhitepaper21,NvidiaWhitepaper20,IntelWhitepaper20} in the deployment of deep neural networks. Using less bits for weights and activations reduces power consumption and increases throughput. However, accuracy is lost, increasingly so for lower bit-widths. This is mitigated by applying an optimizing \textit{quantization method} \cite{Gholami21} for creating the set of HW-consistent parameters, e.g. integer weights and others. It is standard to divide quantization methods into two broad groups – post-training quantization (PTQ) and quantization-aware training (QAT). Broadly speaking, QAT methods involve at least a few epochs of quantization-augmented training on the original labeled train-set while PTQ methods are fast and require a small amount of unlabeled data. PTQ methods are also valued for their robustness and ease of use, e.g. minimal if any per-network handcrafted settings. This is of special interest to HW vendors designing AI accelerators. Their hardware-tailored quantization tools try to provide an automated compilation of quantized network from a pre-trained one, within a fast act that is practical to iterate across deployment settings (e.g., compression levels).

Curiously, it seems there's not much published about how well QAT methods do in the PTQ regime – i.e., when they are constrained to using a small amount of compute resources and unlabeled data, but otherwise adhering to end-to-end weights training. For conciseness, we term this stripped-down QAT the Quantization-aware Finetuning (QFT) regime. In this paper, we explore the potential of this setting to provide a simple but SoTA PTQ baseline while using a small amount of data, no labels, and a quick single-GPU run.

We emphasize joint training of all deployment parameters, including auxiliary ones such as re-coding factors and scales, bringing under that umbrella also the cross-layer factorization \cite{MellerEqual19,NagelDFQ19} formerly treated as pre-quantization conditioning.
To that end, we express HW constraints and relations among parameters as an \textit{offline subgraph} fed by an \textit{independent subset} of DoF, to be cast as variables in a DL framework. In section 3, we lay a foundation to that in terms of a principled HW-aware analysis of a typical arithmetic pipeline, also called for by deployability concerns \cite{LiMQBench21} often arising for works in this field. Curiously, even the standard layerwise/channelwise mode distinction can yet benefit from a rigorous HW-anchored definition, which we supply, and then proceed to uncover and analyze a \textit{doubly-channelwise} kernel quantization mode.
In section 4, we provide extensive experimentation to show that thanks to proper utilization of all DoF, our simple QFT method is on-par with SoTA PTQ methods while having less steps and hand-crafted decisions.
\textbf{Our contributions are as follows:}
\begin{itemize}
    \item We show that QAT scaled down to the PTQ regime can give results on-par with SoTA method when all degrees of freedom (DoF) are jointly finetuned.
    \item We outline a roadmap towards such \textit{fully trainable deployment adaption} for any computational graph. Applied to convolutions deployed on a typical AI accelerator, it generalizes the layerwise/channelwise language to a unified \textit{vector scales} semantics, unlocking a trainable cross-layer factorization.
\end{itemize}

\section{Prior work}

\subsubsection{Quantization aware training (QAT):} 

Straight-through Estimator (STE) was introduced in \cite{BengioSTE13} as a generic enabler of backpropagation through non-differentiable ops, then applied to trained binarization \cite{Hubara16} and integer quantization \cite{Hubara17,Jacob17,Krishna} by injecting round\&clip into the training graph. In \cite{FAQ18} QAT achieving 4-bit accuracy matching or exceeding full-precision was demonstrated. Alternative differentiability solutions were suggested by \cite{LiuAlphablend19,Gong19DSQ,Louizos19,Choi20sparsereg,Fan20extreme} among others, but STE still provides a strong baseline. The training can gain from re-optimizing the quantization grid parameters (e.g. step size) after each epoch \cite{LiuAlphablend19,Sakr22QATclip} or better, extending STE to train them jointly with weights and biases \cite{ChoiPact18,EsserLSQ20,JainTTQ20,Uhlich20}. This enabled better results and/or shorter training \cite{JainTTQ20} which however still entails at least a few epochs on the full labeled training set. On top of that, accurate QAT setup requires careful adherence to both original training procedure and the deployment graph, conflicting demands in case of BatchNorm \cite{Ioffe15} folding, leading to subtle effects \cite{Krishna,Banner18,LiMQBench21}.

\subsubsection{Post-training quantization (PTQ):} 

Many architectures, e.g. ResNets \cite{He16} can quantize well to 8-bit by a trivial round-to-nearest with the grid range set to a naive \(max(|.|)\) applied on weights and data samples (calibration set). In 4-bit weights case, clipping is introduced lest the rounding errors be catastrophic (see Appendix \ref{section:CLE}), typically balancing the error types in a \textsc{mmse} (minimum mean-square error) sense \cite{BannerACIQ18,LiuAlphablend19}. Some architectures s.a. MobileNets \cite{Howard17} still presented a challenge for 8b quantization especially in \textit{layerwise} setting. This was solved by heuristics utilizing two other degrees of freedom. First, a bias correction \cite{FinkelIBC19,NagelDFQ19} zeroing quantization error's 1st moment. Second, the inverse-proportional cross-layer factorization \cite{MellerEqual19,NagelDFQ19}, an equivalence transformation of network weights providing partial \textit{equalization} of channels' ranges. Wide-range channels can also be 'tamed' by splitting \cite{Zhao2019OCS}, at the expense of network signature change and mild resource increase.

While PTQ originally \cite{Jacob17,Krishna} implied no training, described in contrast to QAT, subsequent works partially relaxed that self-limitation. Massively downscaled QAT of a parameter subset was first proposed in \cite{FinkelIBC19} for biases and in \cite{Choukroun19} for multiplicative corrections. In \cite{HubaraAdaquant20,NagelAdaround20,WangBitSplit20,hubara2021accurate} the teacher-student ("reconstruction") training is performed layer-by-layer, avoiding concerns of overfitting at the expense of using a local proxy for the loss. In \cite{NagelAdaround20}, the STE is replaced by a specially designed constraints relaxation procedure, progressively penalizing deviation of weights from grid. In \cite{LiBRECQ21} a block-by-block reconstruction was adopted, better approximating the network loss, at the expense of some block-structure hyperparameters to be manually set per network.

\section{Methods}

\subsection{Quantization-aware Finetuning (QFT) - a downscaled QAT}
We begin by stripping down the standard STE-based Quantization-aware Training to meet PTQ regime, as applied in blind-optimized deployment to accelerators. Specifically, we aim at a method that can accept a pretrained network exported for an \textit{execution runtime} (e.g., \textit{tflite/onnx} format), stripped of any mention of the task, training procedure and dataset, perhaps with some compression already applied (e.g. BatchNorm folding). Ideally, the method runs in a black-box mode, without any per-net hyperparameter tuning by the user.
To that end, we reduce the number of images used for training by 2-3 orders of magnitude, and avoid any use of labels. Instead, we rely on a pure knowledge-distillation (KD) \cite{HintonKD15} method, with the full-precision pre-trained network in \textit{teacher}'s role \cite{MishraMarrQKD18},\cite{PolinoQKD18},\cite{ZhuangQKD18}. The student to be trained is the fake-quantized network, or more precisely, a deployment-aware graph simulating both the \textit{online} (HW run-time) and \textit{offline} (compile-time) computations. 

For the training loss, the classic KD \cite{HintonKD15} cross-entropy loss on logits was found to give inferior results in our small-data regime. Large improvement is achieved by usage of internal layers \cite{ZhuangQKD18}, taking the norm of difference of \textit{teacher}'s and \textit{student}'s (properly decoded) activations. The last layer of backbone alone is a strong baseline, able to supervise all convolutions and making use of spatially-rich distillation signal available before the global pooling. Note that this choice makes the method task-agnostic, in essence perfecting for deployment the feature-extracting backbone, even as we experiment with ImageNet classifiers as the standard benchmark for quantization techniques. See ablation studies in Experiments section for exploration of the design space hereby outlined.

\subsection{What else can be finetuned beyond weights and biases?}
\subsubsection{Quantized deployment parameterization, at its most general.}

The simple finetuning of biases and 4b-quantized weights (using constant \textsc{mmse}-optimal ranges) gives surprisingly robust results. Encouraged by this fact, we seek further improvement by exploiting gradient-driven modifications to *all* parameters of a given deployment. A manifestly optimal extension in that respect is inevitably specific to hardware implementation and network architecture. 
\begin{wrapfigure}{r}{0.5\textwidth}
%\begin{figure}
\vspace{-0.8cm}
\includegraphics[scale=0.7]{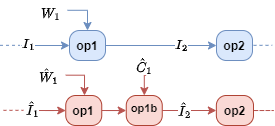}
\caption{Problem abstraction as twin computational graphs - a full precision (FP) one (blue) and a HW-deployed quantized one (red). The latter includes replicas of nodes of the former, edges carrying \textit{encoded} data, and auxiliary \textit{recode} nodes \& constants}
\label{fig:fp_vs_deploy}
\vspace{-0.6cm}
%\end{figure}
\end{wrapfigure}
This is evident in the general problem statement outlined in Fig. \ref{fig:fp_vs_deploy}. We assume for simplicity a \textit{linearly encoded, uniformly quantized} relation between deployed and full-precision tensors: \[ \hat{T} \simeq clip\left(\lfloor T / S_T\rceil\right) + Z_T\ \ \forall{T}\in\{I,W\}\] holding exactly for all pre-trained \textit{weights} (W) and approximately for some \textit{intermediates} (I). Here, \(S_T,Z_T\) are the \textit{scale} and \textit{zero-point} for \(T\), themselves possibly multidimensional tensors reflecting encoding granularity. Now, crucially, some constraints apply, fully determined by the HW implementation and in particular by the properties of the auxiliary nodes (e.g. linearity) and auxiliary constants (e.g. dimensionality). The resultant constraints restrict the dimensions and create relations among the full quantization-parameters set \(\{S,Z,C\}\). Those are to be upheld in the course of any optimization process so to maintain deploy-ability. It is then becomes useful to identify the subset of \textit{degrees of freedom}, the minimal sufficient parameterization of the manifold of graphs both deployable and approximately equivalent to the FP pre-trained network.

\subsubsection{Existing simulation and training approaches} tend to avoid HW-anchored analysis and prefer to abstract the HW away as isolated applications of quantization op. That is, a stand-alone operation to be \textit{injected} into otherwise full-precision graph, e.g., Eq. (\ref{eq:simle_scale_and_quant}) for symmetric-range quantized weights:
\begin{equation}
    Q_{b,s}(x)=s*clip\left(\lfloor x/s\rceil, \pm(2^{b-1}-1)\right)
    \label{eq:simle_scale_and_quant}
\end{equation} 
Here the scale \(s\) is the quantization bin/step size for a certain tensor or slice thereof, hence \textit{per-channel} or \textit{per-tensor} quantization. With the standard application of STE \cite{BengioSTE13} as \textsc{fakequant} \cite{Krishna} the gradient w.r.t \(x\) is taken as 1 (0) inside (outside) of range. In \cite{ChoiPact18,EsserLSQ20,JainTTQ20,Uhlich20} this is extended by an explicit gradient w.r.t \(s\), the scale parameter, which controls the precision/range trade-off, i.e. the amount of \textit{clipping}. Trainable \(s\) provides a partial answer to the title of this section, at least for low-bit channelwise-quantized convolutional kernels.
\subsubsection{The cross-layer inverse-proportional factorization} provides an example of a degree of freedom (DoF) not playing well with the existing approach. This DoF is exploited in \cite{MellerEqual19,NagelDFQ19} to improve \textit{8b, layerwise} PTQ so to approach \textit{channelwise} accuracy. As we show in Appendix \ref{section:CLE}, any fruitful exploitation of this DoF for 4b weights has inevitable interplay with \textit{clipping}, motivating making it trainable similarly to scales. However, until now, this DoF was analyzed \cite{MellerEqual19,NagelDFQ19} as a pre-quantization conditioning, with factors applied to kernel rows and inverse factors applied to columns of previous layer’s kernels. This view provides for a modular tool but on the other hand leaves untapped potential for holistic co-optimization with other DoFs. Below we re-frame the factorization as vector scale parameter, rather than modification to the weights themselves, for a unified and differentiable formulation of \textit{clipping} and \textit{equalization}, providing for full trainability.

While trainable factorization could be arranged for in a targeted fashion, we aim for a more generic analytic recipe that will not only unlock eventual trainability for any and all DoF but also facilitate their initial discovery and mapping. The trainable cross-layer factors DoF will then automatically emerge from a principled analysis, alongside other applications we briefly outline.

\subsection{DoF mapping via \textit{over-parameterized scale tensors}}

\paragraph{We propose the following HW-aware analysis and simulation recipe}:
\begin{enumerate}
    \item Start with over-parameterized scales, i.e. \(S_T\) shaped as the respective \(T\).
    \item Analyze constraints of actual arithmetic pipeline to arrive at an under-determined \textit{system of equations} describing the relations of scales.
    \item Solve for the degrees of freedom (DoF) - a subset of \textit{independent} variables, from which all quantization parameterization can be inferred.
\end{enumerate}
    
In section \ref{section:offline_graph} we show all-DoF \textit{trainability}  to follow from step (3), via differentiable inference of all deployment parameters from the DoF. Here we focus on the DoF \textit{mapping} which we now exemplify for simple bias-less convolutions. 

\begin{figure}
\vspace{-0.5cm}
\centering
    \includegraphics[scale=0.29]{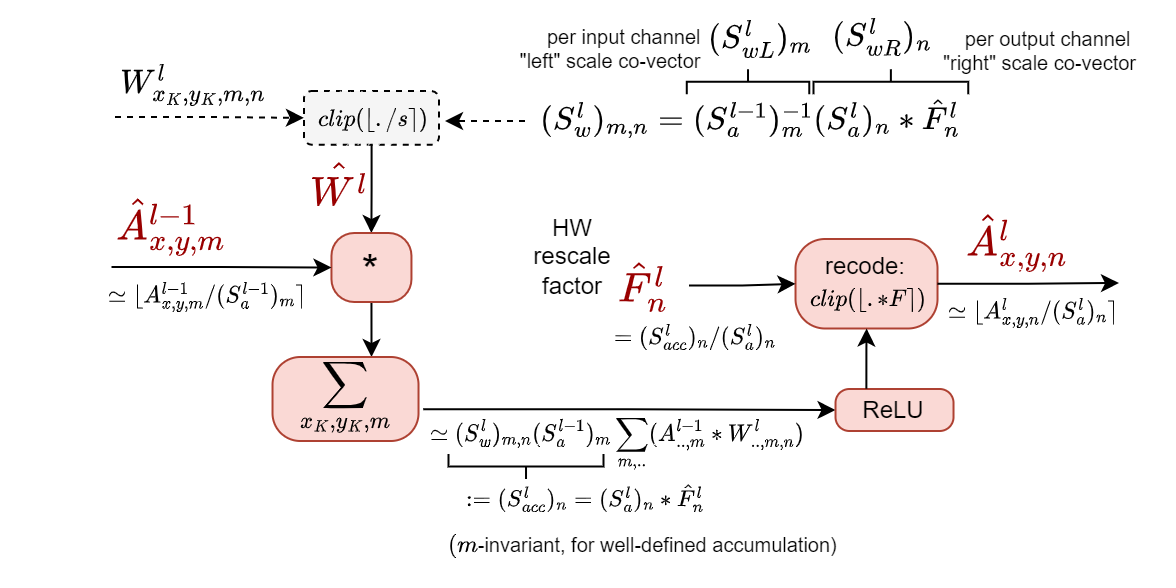}
    \caption{Analyzing quantization parameters constraints and relations for a simple convolution. Can parameterize all via either \(\{S_a,\hat{F}\}\) or \(\{S_{wL}, S_{wR}\}\), at most two vector DoF, less if additional HW constraints apply (layerwise rescale, special activation, etc.).}
\label{fig:vector_scales_single_conv}
\vspace{-0.2cm}
\end{figure}

\subsubsection{Degrees-of-freedom analysis for a simple convolution sequence}
\label{section:DoFConvAnalysis}
In Fig. \ref{fig:vector_scales_single_conv} we present an exhaustive analysis of single convolution's fully-integer deployment arithmetic, assuming for brevity of exposition a symmetric encoding, and no bias. Quantized weights \(\hat{W}\) are derived offline from full-precision ones \(W\) by scaling with an over-parameterized scale tensor \((S_w)_{m,n}\)\footnote{Given a tacit assumption of HW-configuration invariance across spatial coordinates.} and round\&clip op. Quantized activations \(\hat{A}\) are in approximate correspondence to activations in a full-precision run, via \textbf{per-channel (vector) scale \((S_a)^l_m\)}. Activation function is followed by an auxiliary \textit{recode} operation, for a symmetric case comprising just a multiplication by factor \(\hat{F_n}\). Note that such op always exists, even if implicitly as a \textit{dequantize+requantize} for non-fully-integer platforms restoring full FP32 representation between layers.
This enables a rigorous definition of per-channel/per-layer hardware specification as the rank of the HW tensor \(\hat{F_n}  \):
% \begin{itemize}
%     \item \textbf{Layerwise}: \textit{scalar}, \(n\)-independent \( \hat{F_n}=\hat{F_0}\mspace{4mu} \forall{n} \).
%     \item \textbf{Channelwise}: \textit{vector}, \(n\)-dependent \(\hat{F_n}  \)
% \end{itemize}
\begin{align*}
&\textbf{layerwise}: scalar\ \hat{F_n}=\hat{F_0}\ \forall{n}.\\
&\textbf{channelwise}: vector\ \hat{F_n}
\end{align*}
Now, as we apply the two constraints of linear computational elements\footnote{Assuming scale-invariant, homogeneous (i.e., \textit{[l]ReLU}) activation function}: (a) All terms in the partial sum having the same scale (b) Scale change by constant upon multiplicative recode, we arrive\footnote{Proof is sketched within Fig. \ref{fig:vector_scales_single_conv} and fully detailed in Appendix \ref{section:ProofOfLR}} at the following relations:
\begin{equation}
\begin{split}
(S^{l}_{w})_{m,n} := & (S^l_{wL})_m * (S^l_{wR})_n \\
(S^l_{wL})_m = & (S^{l-1}_a)_m^{-1}  \\
(S^l_{wR})_n = & (S^l_a)_n*\hat{F}^l_n
\end{split}
\label{eq:LRfromAF}
\end{equation}
These restricts the kernel scale matrix to an outer product of \textit{left} and \textit{right} co-vectors\footnote{Curiously, the two "scale co-vectors" are not directly interpretable as "quantization step size" of a certain tensor slice, exposing the typical abstraction (Eq. (\ref{eq:simle_scale_and_quant})) as an oversimplification for even the standard uniform quantization.}. Eqs.\ref{eq:LRfromAF} rigorously map and define the vector (per-channel) degrees of freedom for a sequence of simple convolutions, with one (two) such DoF for layerwise (channelwise) HW rescale capability.

\paragraph{Corollary 1: The cross-layer factorization (CLE) \cite{MellerEqual19,NagelDFQ19} vector DoF}. It emerges immediately from Eqs.\ref{eq:LRfromAF} as the freedom to change \((S^l_a)_m\). In \textit{layerwise} setup, it's the single vector DoF, fully determining,  up to a scalar, both \((S^l_{wL})_m, (S^{l-1}_{wR})_n\), the left (right) scale of the following (preceding) layers' kernels. This is equivalent to CLE factors (see Appendix \ref{section:CLE}) in the sense of the overall relation between the FP-trained and quantized kernels. However, here this relation is viewed as part and parcel of quantization instead of as equivalence transform of fully-precision network; emergent from principled mapping not special insight. This approach facilitates further analysis of less trivial cases, e.g. fan-out from \(l-1\) to multiple layers \(l_j\) which then share \((S^l_a)_m\) and their derived \((S^{l_j}_{wL})_m\). 

\paragraph{Corollary 2: Possibility of \textbf{doubly-channelwise} kernel quantization.}
For a \textit{channelwise} HW setup, the freedom to choose re-coding factors \(\hat{F}^l_n\) of same layer fully determines the right kernel scales \((S^l_{wR})_n\); this is equivalent to and well described by the usual abstracted-HW treatment. However, absent from existing discourse, to our best knowledge, is the combo with the CLE DoF discussed above, yielding a \textit{doubly-channelwise} quantization of kernel, now scaled with both per-input-channel and per-output-channel granularity. To analyze such a scheme, we invert Eq. (\ref{eq:LRfromAF}) with the kernel scale co-vectors as independent variables, determining the rest of quantization parameters:

\begin{equation} 
    (S^{l-1}_a)_m = (S^l_{wL})_m^{-1}\ \ ; \quad  (S^{l}_a)_n = (S^{l+1}_{wL})_n^{-1}
\label{eq:LR_reparam1}
\end{equation}

\begin{equation} 
    \hat{F}^l_n = (S^l_{wR})_n / (S^l_a)_n = (S^l_{wR})_n  (S^{l+1}_{wL})_n
    \label{eq:LR_reparam2}
\end{equation}

%\begin{figure*}
\begin{wrapfigure}{r}{0.5\textwidth}
    %[!hb]\centering
    \vspace{-0.7cm}
    \includegraphics[scale=0.2]{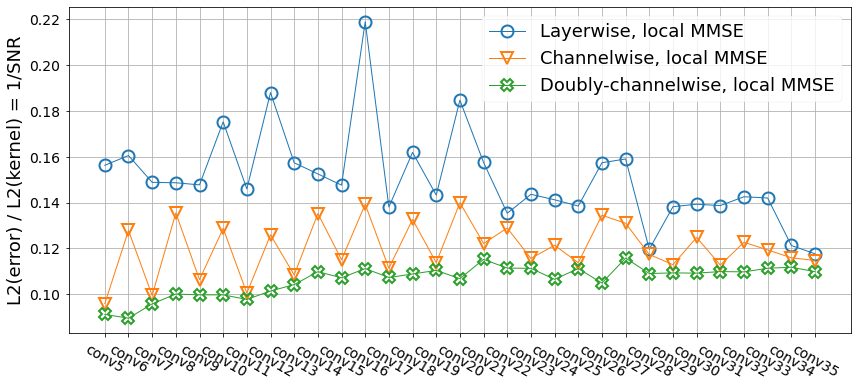}
    \caption{MobileNetV2 kernels quantization error norm, across scale-tensor granularity. For this net there seems to be gain from every extra vector degree of freedom, at least in terms of local error reduction.} %. We optimize using 
    \label{fig:quantization_error}
    \vspace{-0.3cm}
%\end{figure*}
\end{wrapfigure}

This scheme requires per-channel and \textit{per-output-edge} recode factors \(\hat{F}^{l-1}_{n,j}\). Another caveat is \((S^l_a)_n\) fully determined (up to scalar) by kernel-optimizing \((S^{l+1}_{wL})_n\), but for W4A8 setting the focus on weights is mostly beneficial. As a basic test of the error-reduction potential, we experiment with local-\textsc{mmse} (minimum mean-square error) quantization of convolutional kernels, see Fig. \ref{fig:quantization_error}. Layerwise, channelwise, doubly-channelwise (\textit{dCh}) \textsc{mmse} problems are defined as below. % (\(||.||\) standing for L2 norm: 

\begin{equation}
\resizebox{0.85\textwidth}{!}{$%
\begin{split} 
& MMSE(W):=\min_{s}\left\Vert W - s*clip\left(\left\lfloor W/s\right\rceil\right)\right\Vert \\
& MMSE_{Ch}(W) := \underset{S_{wR}}{\min}\left\Vert W_{m,n} - (S_{wR})_n*clip\left(\left\lfloor\frac{W_{m,n}}{(S_{wR})_n}\right\rceil\right)\right\Vert = \sqrt{\sum_nMMSE^2(W_{n,..})} \\
& MMSE_{dCh}(W) := \underset{S_{wL}, S_{wR}}{\min}\left\Vert W_{m,n} - (S_{wL})_m(S_{wR})_n*clip\left(\left\lfloor\frac{W_{m,n}}{(S_{wL})_m(S_{wR})_n}\right\rceil\right)\right\Vert
\end{split}$%
}
\label{eq:MMSE}
\end{equation}

We solve for the \textsc{mmse} using \textit{PPQ} algorithm adopted from \cite{LiuAlphablend19}, which we extend for the inseparable doubly-channelwise case by a novel procedure using alternating rows/columns projections (Appendix \ref{section:APQ}). 

\subsection{Making all DoF trainable via the \textit{offline subgraph}}
\label{section:offline_graph}
\begin{figure}
    \centering
    \vspace{-0.5cm}
    \includegraphics[scale=0.35]{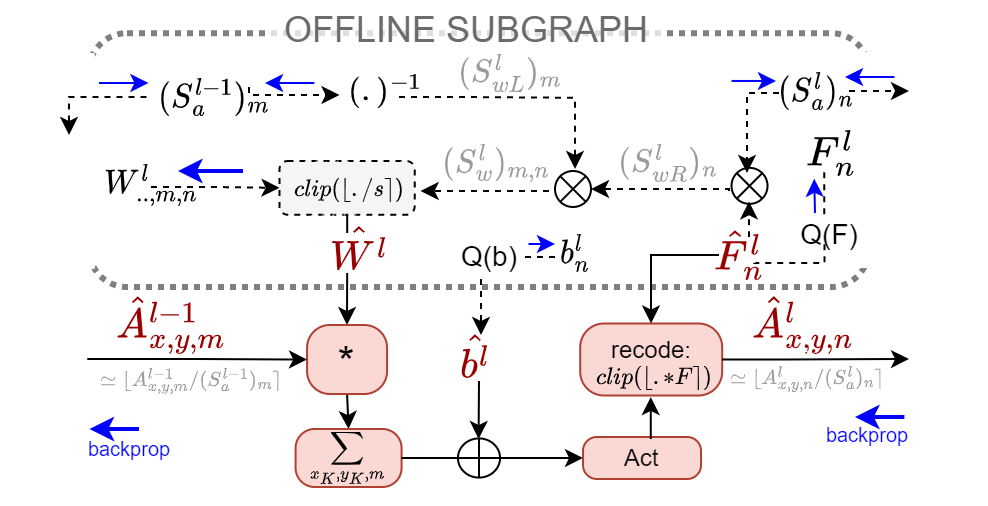}
    \caption{Exhaustive elaboration of the \textit{offline subgraph} for convolutions sequence. Layerwise quantization setting is supported by demoting the rescale factor to scalar \(\hat{F^l}\). The output of the offline subgraph is the \textit{exports} \(\hat{W},\hat{F}\) fed into the online, HW-emulating part. The combined graph is end-to-end differentiable by virtue of Straight-through Estimator (STE) \cite{BengioSTE13} applied to any \(clip(\lfloor.\rceil)\) op, representing the lossy part of the online/offline quantization of activations/weights, respectively. This naturally extends to any additional HW-specific quant-op, e.g. Q(b), Q(F). Weights, biases and scaling freedoms - all endpoints of the backprop path (blue) - can be end-to-end trained on the same footing, while train-deploy consistency is maintained by enforcing all constraints in the forward pass of the offline subgraph.}
    \label{fig:offline_subgraph}
    \vspace{-0.5cm}
\end{figure}

Central to our analysis is the overparameterized quantization and its resolution from degrees-of-freedom (DoF). With the HW constraints and relations among the quantization parameters described by a system of equations, we define the \textit{offline subgraph} as its formal solution. It amounts to a computation inferring all deployment and auxiliary constants (e.g. quantized weights, scales, etc) from their maximal unconstrained subset, hence the DoF set. For a simple convolutions sequence we analyze in detail here, the offline subgraph can amount to applying Eq. (\ref{eq:LRfromAF}) to the DoF set of weights, activations' scales and rescale factors: 
\begin{equation}
{trainables}\ :=\ \{W^l_{..,m,n}, (S_A^l)_n, \hat{F^l_n}\ |\ l=1..L, n=1..N_l, m=1..N_{l-1}\}
\label{eq:vars2train}
\end{equation}
This is a convenient choice of independently trainable subset of quantization parameters. The full simulation graph is depicted in Fig. \ref{fig:offline_subgraph}, explicitly delineating the \textit{online}, HW-runtime emulating part and the \textit{offline}, compile-time part.

This approach replaces explicit quantization-parameter gradient definitions \cite{ChoiPact18,EsserLSQ20,JainTTQ20} by a \textit{native} gradient flow through the offline subgraph. This enables for the first time an end-to-end training of all quantization parameters that are interlinked or not localized to a single Eq.(\ref{eq:simle_scale_and_quant})-style expression as implied by LSQ \cite{EsserLSQ20} and similar methods. This includes the cross-layer factorization vector DoF, now represented as a fully-trainable activation-scale variable.

For application of the offline-graph concept in real deployments, the equations have to be generalized in a few directions which we broadly outline:
\begin{enumerate}
    \item Additive arithmetic (bias, ew-add), \textit{zero-point} parameters joining the scales for asymmetric encoding, and their respective relations.
    \item Other layer types - e.g. \textit{depthwise} convs not having L/R scale but only one, \textit{elementwise} add/multiply introducing more sets of rescale-factors, \textit{non-arithmetic} layers s.a. Maxpool,Concat,ResizeNN introducing non-parametric relations between vector scales of their input and output, etc.
    \item More specialized constraints - quantization of rescale-factors, non-homogeneous activation functions precluding cross-layer factorization, etc.
\end{enumerate}
These are similarly achieved by repeated application of same 2 principles: (A) Explicitly model all HW arithmetic in the online subgraph (B) Express all constraints resolution in the offline subgraph. See Appendix \ref{section: rethink simulation} for further detail.

% The benefit of our HW-centric view of arithmetic operating on lossy-encoded I/O (Appendix \ref{section: rethink simulation}) is fully unlocked when systematically extended for the "long tail" of layer types and HW idiosyncrasy. 

%\pagebreak
\section{Experiments}

We apply our all-DoF QFT to standard ImageNet-1K \cite{ImageNet15} classification CNNs \cite{ResNet16,RegNet20,MnasNet19,MobilenetV2_18}, pre-trained courtesy of \cite{LiBRECQ21}. BatchNorm ops are folded back into convs. We use in-house simulation library wrapping \textit{keras, tensorflow} \cite{chollet2015keras,Abadi16}.

We experiment with two 4b-weight setups. In the 'permissive' one, we use per-channel rescale factors, and weight-only quantization; we utilize the doubly-channelwise quantization, training two vector DoF per layer. In the 'deployment-oriented' one, we use 8b (unsigned) activations and layerwise rescale factors; here, only the cross-layer activation vector scale DoF is available to be finetuned jointly with weights and biases.
While it is common to quantize all layers except the first (which is left in full precision), we note following \cite{Gluska20} that this is quite an arbitrary choice. Instead, for a flat overhead rate across nets, we quantize in 8b a few smallest layers, added-up by increasing size till their cumulative weight-memory footprint is 1\% of the total across the convolutional backbone. Impact on resource and deployment is equally negligible. In a real HW deployment scenario more sophisticated heterogeneous quantization would be applied, while on the other hand, additional lossy elements apply (e.g. quantization of bias/accumulator, rescale factors, activation, etc.), and further compression (4b-activations, pruning) is considered. Here we focus on the challenge of accurate 99\%-4b-weights backbone within a simple single-stage method.

The hyperparameters of the Quantization-aware Finetuning are uniform across the networks and experiments: Distillation from the FP net serving as teacher, using a normalized L2 \textit{reconstruction} loss at the input to the average-pooling layer (equivalently, output of the convolutional backbone). We train for 12 epochs of 8K images each, with batch size of 16, \textit{adam} optimizer and cosine learning rate schedule, decaying across 4 epochs starting from 1e-4 and reloading at /2 (i.e. 5e-5, 2.5e-5 @ epoch=4,8), without any regularization or augmentation. Quantization is initialized by \textit{scalar} (per-tensor) scales derived from naive (\textit{max-min}) range calibration for activations and \textit{mmse} (Eq. (\ref{eq:MMSE}a)) for weights, then used to compute \(F^l\) via inversion of Eq. (\ref{eq:LRfromAF}); a sole pre-QFT step.

\begin{figure}
\vspace{-0.7cm}
\includegraphics[scale=0.22]{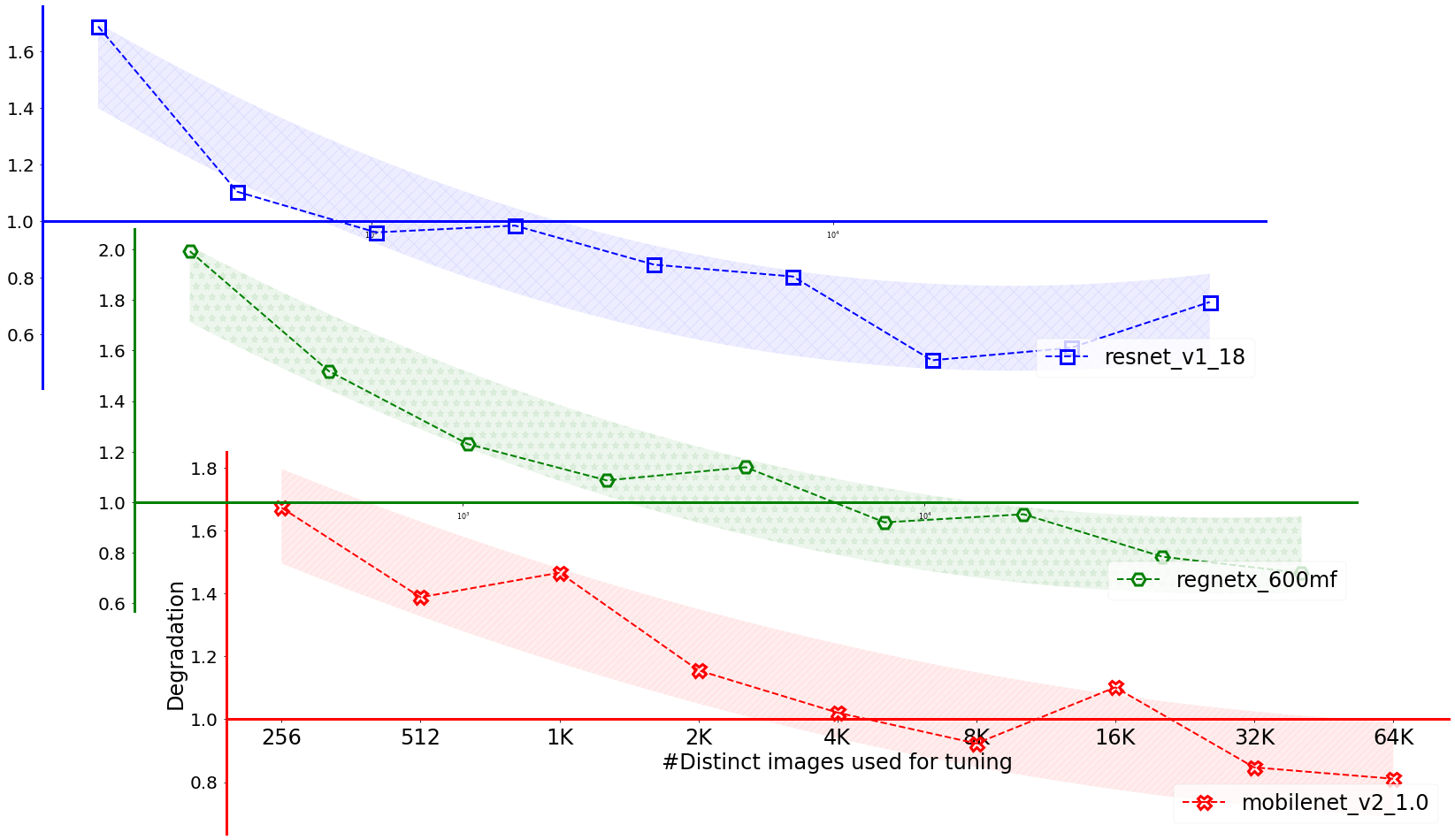}
\caption{Effect of dataset size on QFT accuracy restoration capability. Overall trend and uncertainty are visualized with \(\pm2\sigma\) bands around a polynomial fit curve.}
\label{fig:data_size_ablation}
\vspace{-0.7cm}
\end{figure}
%\end{wrapfigure}

\begin{wrapfigure}{r}{0.5\textwidth}
%\begin{figure}
\vspace{-1.8cm}
\includegraphics[scale=0.25]{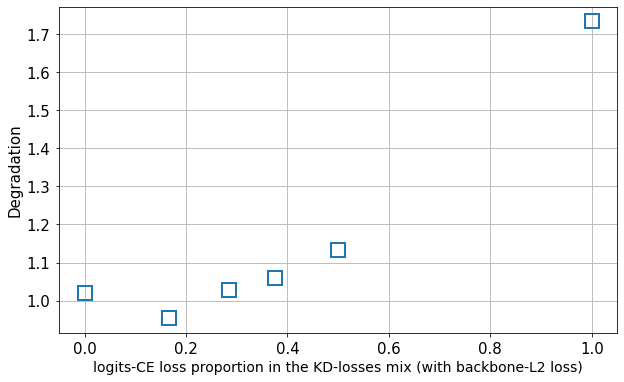}
\caption{Effect of complex KD loss - mixing in cross-entropy on logits (\textit{mobilenet v2})}
\label{fig:CE_vs_L2}
\vspace{-0.6cm}
%\end{figure}
\end{wrapfigure}

\subsection{Ablation studies}
% \begin{wrapfigure}{r}{0.5\textwidth}

\subsubsection{Dataset size ablation} is presented in Fig. \ref{fig:data_size_ablation}. We scale down distinct images used but increase epochs to keep the total images fed constant at 32K, so convergence properties are roughly the same. We see that results deteriorate quite gracefully down to 1K images and well below, not displaying a strong overfitting effect in contrast to concerns voiced in \cite{HubaraAdaquant20},\cite{LiBRECQ21}. Beyond a few K images there seem to be diminishing returns at least for our "safe" LR=1e-4 regime. We therefore set 8K images (~0.7\% of the training set) as our working point for the rest of the section. 

\subsubsection{QFT hyperparameters ablation}
QFT easily supports more complex knowledge-distillation (KD) losses. Here we experiment with adding the classic \cite{HintonKD15} cross-entropy (CE) loss at the logits layer, mixed-in in varying proportion (weighting) with our default backbone-output L2 loss. Results for one network in which some benefit was found for low CE proportion are depicted in Fig. \ref{fig:CE_vs_L2}. 
\begin{wrapfigure}{r}{0.5\textwidth}
%\begin{figure}
\vspace{-0.7cm}
\includegraphics[scale=0.25]{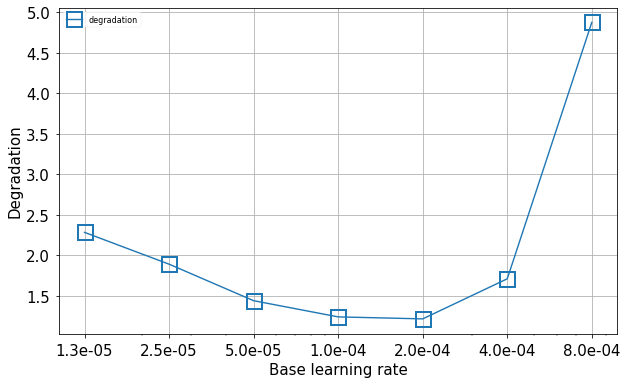}
\caption{Effect of base LR (\textit{regnetx 600mf})}
\label{fig:LR_ablate}
\vspace{-0.5cm}
%\end{figure}
\end{wrapfigure}
The effect is by and large detrimental, e.g. when using CE-logits alone (proportion=1.0) the resultant degradation is almost double, and even greater in other nets we tested. Usage of CE-logits as default can partially explain weak results reported on end-to-end finetuning by previous works \cite{HubaraAdaquant20}, driving the trend towards layer-by-layer training which we challenge in this work. Another important hyperparameter is the (base) learning rate, which we explore for one network in Fig. \ref{fig:LR_ablate}. There seem to be a robust performance region around 1e-4; this seems to carry over well to all nets.

\begin{figure}
\vspace{-0.2cm}
\includegraphics[scale=0.25]{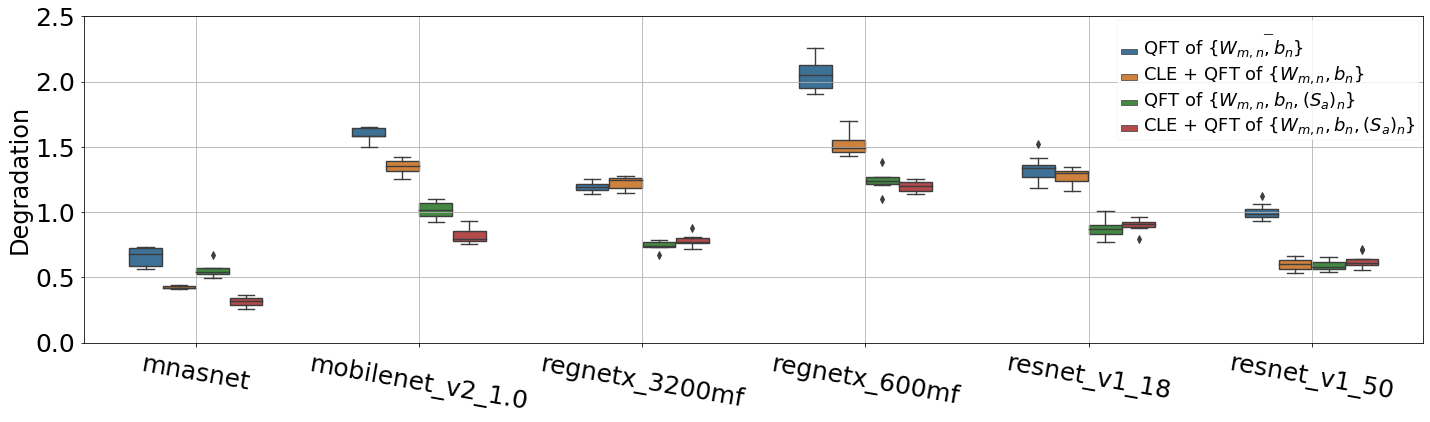}
\caption{Layerwise quantization (8bA, 4bW) with QFT: Optimizing the cross-layer factorization DoF by applying 4b-adapted CLE heuristic before QFT \textbf{and/or} by training this DoF within the QFT (see Fig. \ref{fig:offline_subgraph} for method detail). We can see that pre-QFT application of CLE (yellow) may yield improvement  over the baseline (blue) of ignoring the DoF, but typically the jointly-trained end-to-end QFT (green) gives smaller degradation. Synergy of the methods (red) is observed for \textit{mnasnet},\textit{mobilenet}, with the CLE serving as a better pre-QFT initialization of this DoF.}
\label{fig:CLE_vs_QFT}
\vspace{-0.7cm}
\end{figure}

\subsubsection{Trained vector activation scale vs. Cross-Layer Equalization (CLE) \cite{NagelDFQ19}}
For the layerwise setup, we explore the approaches to utilizing the cross-layer factorization degree-of-freedom (CLF DoF) \cite{MellerEqual19,NagelDFQ19} appearing in our analysis as the activation vector scale \((S^{l}_a)_n\). We use our own adaptation of the CLE method, optimal for 4-bit weights, see Appendix \ref{section:CLE} for details. In a nutshell, we use \textsc{mmse}-optimal row/column range values as inputs to geometric-mean heuristic instead of \(max(|.|)\). For this section's ablation, we initialize QFT with the activation vector scale either modified by the above or left trivially uniform. Then we run QFT with the vector scale either trained along with the weights\&biases or frozen. The accuracy degradation results for the resultant 2x2 configurations can be found in Fig. \ref{fig:CLE_vs_QFT}. The added value of the joint all-DoF finetuning is evident for most networks, with the two-step CLE + all-DoF QFT yielding best results.

\subsubsection{Trained channelwise quantization}
\begin{figure}
\vspace{-0.3cm}
\includegraphics[scale=0.25]{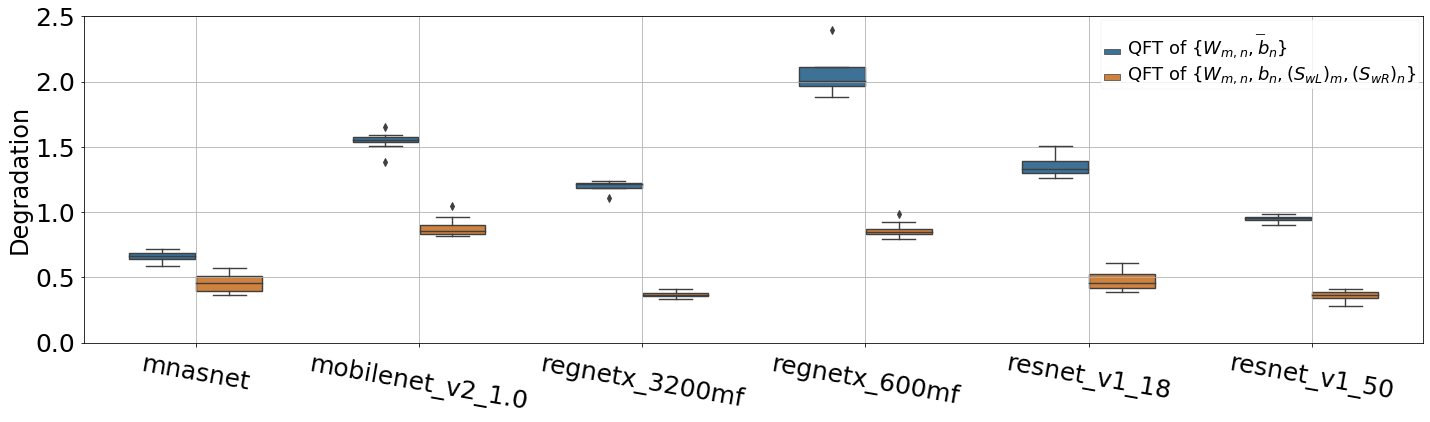}
\caption{Doubly-channelwise (4bW) quantization with QFT: Effect of training \((S_{wL})_m, (S_{wR})_n\) within the finetuning process jointly with weights \& biases}
\label{fig:doubly_QFT}
\vspace{-0.7cm}
\end{figure}

We now turn to the \textit{channelwise} scheme. We assume HW capable of per-channel rescale factors, which together with the CLF DoF unlocks doubly-channelwise kernel quantization (see Methods). As far as DoF parameterization goes, either Eq.(\ref{eq:vars2train}), or the explicit \((S_{WL})_m, (S_{WR})_n\) (Eqs. (\ref{eq:LR_reparam1},\ref{eq:LR_reparam2})) can work for trainable variable designation. We train both vectors together with weights and biases, starting from the plain uniform initialization, for simplicity. Here the results (Fig. \ref{fig:doubly_QFT}) show even greater, up to x3, decrease in degradation thanks to training all vector quantization DoF. Note that the \textit{frozen-scales} result (blue) in Fig. \ref{fig:doubly_QFT} is very similar to Fig. \ref{fig:CLE_vs_QFT}, testifying to the negligible effect of 8b activation quantization in these networks. The gain comes from proper utilization by end-to-end training of the extra vector DoF.

\subsection{Comparison to state-of-the-art methods on ImageNet-1K}
In Table \ref{table:results_brecq_nets}, we compare our results obtained by all-DoF QFT to SoTA PTQ methods, in layerwise and channelwise regime as described above. For the layerwise regime we also present results of 2-step pipeline, with vector-scales training initialized with our 4b-weights adaptation of CLE (Appendix \ref{section:CLE}), achieving further improvement for \textit{mobilenet, mnasnet}.

For the  \textit{4/8,lw} regime we outperform other methods, achieving sub-1.0 degradation for most nets. For the \textit{4/32,chw} case we achieve sub-0.5 degradation for most nets, on-par with SoTA \cite{LiBRECQ21}. Across all QFT experiments we don't apply any per-network parameter or procedure modification. Further gains per net may be had by trying: (A) Block-by-block reconstruction as in \cite{LiBRECQ21} (B) More complex initialization, e.g., Eq. (\ref{eq:DiChMMSE}) (C) Scan for best learning rate (and other hyperparameters) - or any other standard train-tuning technique. We stress that QFT is a single-stage method performing holistic optimization of DoF typically tuned in separate PTQ stages \cite{NagelWhitepaper21} - clipping, equalization, bias-correction, etc.

\begin{table*}[t]
\caption \small{Comparing QFT to SoTA PTQ methods. chw/lw stands for channelwise/layerwise. Results of other methods are quoted from respective papers and compared to respective full-precision evaluations for fair degradation values (in parentheses).
}
\label{table:results_brecq_nets}
% \usepackage{booktabs}
\resizebox{\textwidth}{!}{%
\begin{tabular}{@{}lllllllll@{}}
\toprule
\multicolumn{8}{l}{ImageNet-1K accuracy (\textit{-degradation}); results within 0.1 of best method in category in \textbf{bold}} \\ \midrule
{\textbf{Methods}} &
  \textbf{\scriptsize{Bits (W/A)}} &
  \textbf{ResNet18} &
  \textbf{\scriptsize{MobileNetV2}} &
  \textbf{\scriptsize{RegNet0.6G}} &
  \textbf{MnasNet2} &
  \textbf{ResNet50} &
  \textbf{\scriptsize{RegNet3.2G}} \\ \midrule
Full Precision &
  32/32 &
  71.25 &
  72.8 &
  73.8 &
  76.65 &
  76.8 &
  78.5 \\ \midrule
Adaround \cite{NagelAdaround20} &
  4/32, lw &
  68.7 (-1.0) &
  69.8(-1.9) &
  72.0 (-1.7) &
  74.9 (-1.8) &
  75.2 (-0.9) &
  77.1 (-1.3) \\
Adaround \cite{NagelAdaround20} &
  4/8, lw &
  68.6 \textbf{(-1.1)} &
  69.3 (-2.4) &
  - &
  - &
  75.0 (-1.1) &
  - \\
QFT (ours) &
  4/8, lw &
  70.35 \textbf{(-0.9)} &
  71.8 (-1.0) &
  72.6 \textbf{(-1.2)} &
  76.1 (-0.55) &
  76.2 \textbf{(-0.6)} &
  77.7 \textbf{(-0.8)} \\
CLE+QFT (ours) &
  4/8, lw &
  70.35 \textbf{(-0.9)} &
  72.0 \textbf{(-0.8)} &
  72.6 \textbf{(-1.2)} &
  76.35 \textbf{(-0.3)} &
  76.2 \textbf{(-0.6)} &
  77.7 \textbf{(-0.8)} \\ \midrule
BRECQ \cite{LiBRECQ21} &
  4/32, chw &
  70.7 \textbf{(-0.4)} &
  71.65 \textbf{(-0.85)} &
  73.0 \textbf{(-0.7)} &
  76.0 (-0.7) &
  76.3 (-0.7) &
  78.05 \textbf{(-0.3)} \\ 
QFT (ours) &
  4/32, chw &
  70.8 \textbf{(-0.45)} &
  71.9 \textbf{(-0.9)} &
  73.0 \textbf{(-0.85)} &
  76.2 \textbf{(-0.45)} &
  76.45 \textbf{(-0.35)} &
  78.15\textbf{(-0.35)} \\ 
 \bottomrule
\end{tabular}}
\vspace{-0.3cm}
\end{table*}
QFT is also a very fast and scalable method. Run times for experiments here vary between 10min for \textit{resnet18} to 50min for \textit{regnetx3.2gf}, on a single NVIDIA RTX A4000. Speed can be credited to a standard end-to-end usage of GPUs and SW stack, in contrast to layer-by-layer methods s.a. \cite{NagelAdaround20,HubaraAdaquant20,LiBRECQ21} which at times hit an IO bottleneck, making high GPU utilization a challenge.
\section{Conclusions and Outlook}
In this work we bridge the gap between two long-standing complementary approaches to DNN quantization - PTQ and QAT. We show that a successful PTQ method can be constructed from QAT operating in quick finetuning mode (hence "Quantization-aware Finetuning" - QFT), stripped of labels and big data usage, \textit{batchnorm} quantization issues and any per-network configuration. QFT introduces into the PTQ domain: (a) an end-to-end holistic optimization approach and (b) application of ideas from the QAT literature. We exemplify the latter point by adopting the \textit{trainable scales} concept \cite{ChoiPact18,EsserLSQ20,JainTTQ20,Uhlich20}, which we rigorously generalize towards all-DoF training. That is done through a first-principles rethinking of deployment parameterization which informs both quantization analysis and simulation. We show that QFT can be competitive for 4bW PTQ, while kept simple (plain STE, end-to-end, single-step, etc.) and without many of the bells and whistles of SoTA PTQ methods. Our SoTA-matching results are underpinned by jointly training all degrees-of-freedom of a given deployment. This feature naturally creates a principled way to optimally adapt our method per HW configuration - layerwise, channelwise, etc. We thus suggest QFT as a simple and strong PTQ baseline, further extendable by a scale-up across the PTQ-QAT continuum and adoption of ideas from other methods. Finally, we envision wider applications across NN compression, e.g., in the context of recent trend \cite{yu2020joint,lazarevich2021post,shomron2021post,zhang2021training,frantar2022optimal,park2022quantized} towards post-training joint pruning and quantization. We leave the exploration of these directions to future work.

% \pagebreak

\bibliographystyle{splncs}
\bibliography{refs}

\appendix
\pagebreak
\section{Rethinking quantization simulation}
\label{section: rethink simulation}
\begin{wrapfigure}{r}{0.6\textwidth}
\vspace{-0.2cm}    
    \includegraphics[scale=0.18]{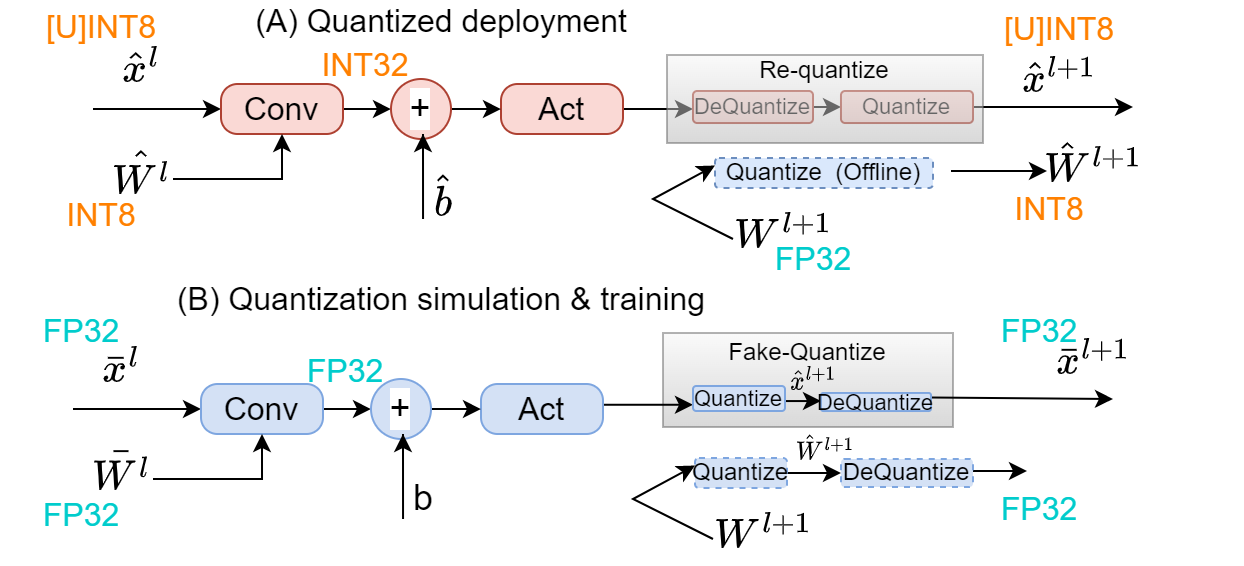}
    \caption{Typical approach for simulation and QAT graph construction, reproduced from \cite{LiMQBench21}, fig.2 there}
\vspace{-0.3cm}    
\label{fig:legacy_simulation}
\end{wrapfigure}

Implicitly underlying our methods is a rethinking of the typical approach to quantization simulation and finetuning, as depicted in  Fig. \ref{fig:legacy_simulation}. The issue with the latter is that the simulation graph follows closely the original, full precision graph - up to injecting \textsc{FakeQuant} ops - rather than the deployment graph. This framework, seemingly a welcome abstraction from HW specifics, in fact encourages a gap between simulation and deployment, recently \cite{LiMQBench21} pointed out as a source of major deploy-ability concerns for many a published method. We argue that an ideal compressed-inference simulator should at least explicitly model the common patterns, such as layer-output rescale factors and their relations to scales. It then should facilitate configurable HW-specific extensions, e.g. layer-output or elementwise-add rescale factors restricted to be powers-of-two.

\begin{figure*}
\vspace{-0.3cm}  
\includegraphics[scale=0.25]{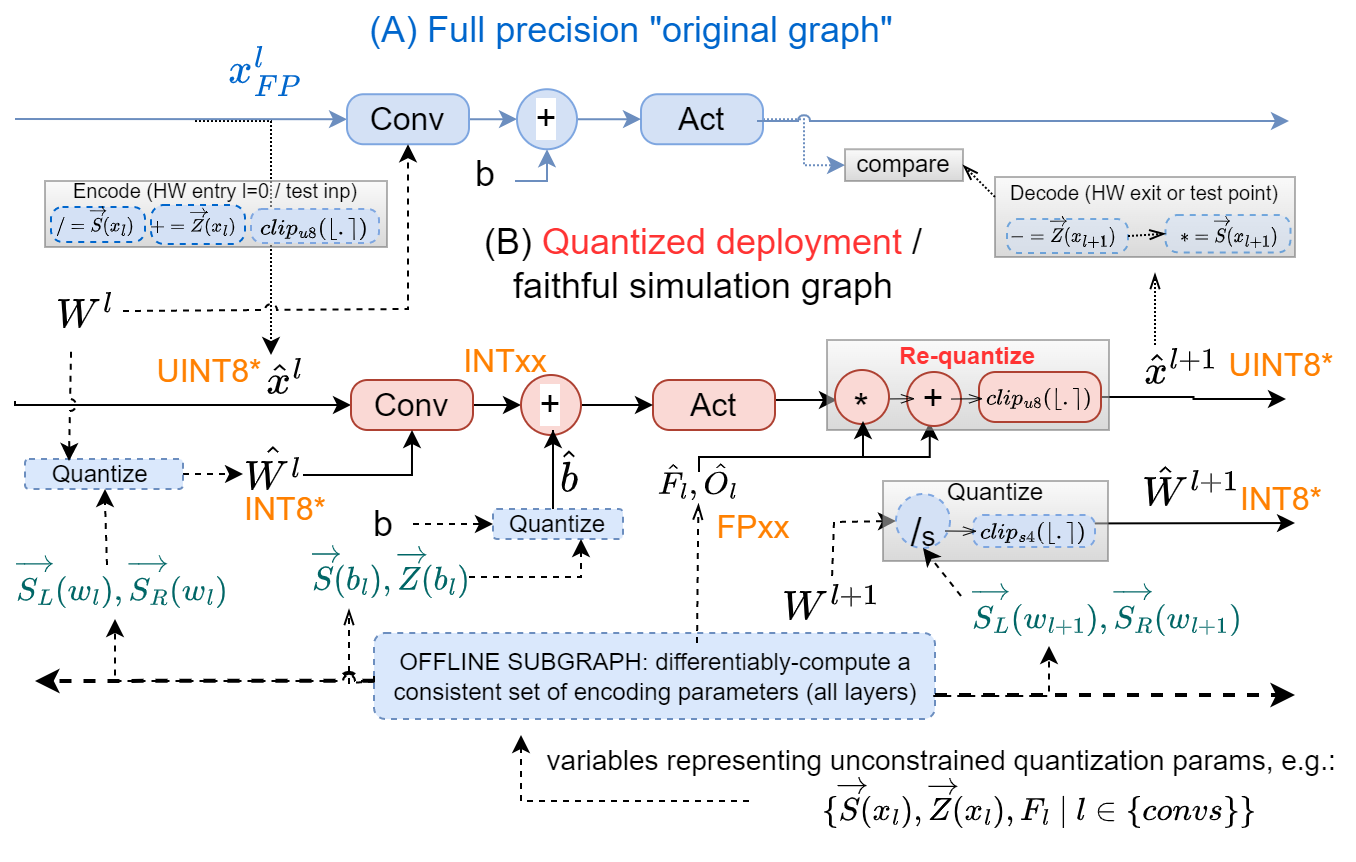} % exported w zoom=200%
\caption{Our suggestion for a HW-aware simulation (\&training) approach. Note slight differences from main text in semantics (e.g., \(x_l\) is input of layer \(l\)), bit-width, etc.}
\label{fig:proposed_simulation}
\vspace{-0.2cm}  
\end{figure*}

Our simulation scheme (Fig. \ref{fig:proposed_simulation}) goes for a maximal sync of the simulation graph to the deployment one (solid red nodes, edges) in its \textit{online} part, and prepends an \textit{offline} subgraph (dashed blue nodes, edges). The online \textit{Requantize} and offline \textit{Quantize} are explicitly modeled and broken down into the arithmetic (e.g., *=scale) and purely bit-discarding (\textit{lossy}) elements. The latter are just \(clip(\lfloor.\rceil)\) ops in the uniform quantization case we focus on here, but easily generalize to nonuniform schemes, for \textit{pruning} by multiplication with a binary mask, etc.
The fake-vs.-real gap is shrunk to an immaterial difference of representation of [u]INTs by FP32 (e.g. [2.0,-3.0,..] sampled from a HW-simulating tensor, hence INT8*), and, for training, a decoration of online and offline bit-discarding elements by Straight-Through Estimator (STE) gradient guide, amounting to \[fakeQuant(x, 0, 2^b-1)\] (for unsigned) in typical DL-framework semantics. Note that this STE application is not burdened with any special provision \cite{EsserLSQ20,ChoiPact18} for training of scales. Their trainability is instead provided for by the fact that STE here is applied on HW-simulating tensor impacted by encoding parameters and ultimately their DoF subset via the \textit{offline subgraph}. Alternative solution for differentiability of bit-discarding ops (s.a. \textit{Adaround}\cite{NagelAdaround20}, \textit{AlphaBlend}\cite{LiuAlphablend19}, etc.) can be easily set up as a drop-in replacement for STE within the \textit{lossy-element} implementation.

The linear relation \(T \simeq S(T)*(\hat{T}-Z(T))\) between each tensor's incarnations \(T, \hat{T}\)  in the FP and simulated-deployment twin graphs, defines a scale \(S(T)\) and a zero-point \(Z(T)\).
These channelwise (or doubly-channelwise, for weights-scale) tensors, together with others such as HW rescale factors and offsets constitute an \textit{overparameterized} set of parameters, to which constraints and relations apply. These are recast as a feed-forward computation, an \textit{offline subgraph} fed by a subset of unconstrained trainable variables representing degrees-of-freedom (DoF), and inferring all encoding parameters and deployment exports (\(\hat{W},\hat{b},\hat{F}\),etc.). The \textit{multiplicative} relations of scales discussed in main text are joined by \textit{additive} relations of zero-points, e.g. for activation-input \(y\), we sum-up the constituent offsets to arrive at:
\begin{equation}
    {Z_n}(y^l) := \sum_mZ_m(x^l)W^l_{m,n}+Z(b_l)
    \label{eq:residue}
\end{equation}
  
Similarly to how the CLE DoF \cite{MellerEqual19,NagelDFQ19} immediately follows from our analytic recipe applied to relations of scales (section \ref{section:DoFConvAnalysis}, Appendix \ref{section:ProofOfLR}), the zero-points' relation in Eq. (\ref{eq:residue}) directly leads to quantized bias absorbing the so-called \textit{residue} \cite{Krishna,LiMQBench21}, a consequence of unsigned representation. That, by setting a constraint \({Z_n}(y^l)=0\) and solving Eq. (\ref{eq:residue}) for \(Z(b_l)\) to get quantized bias as:  \[\hat{b}_n=b_n/S_n(b)+Z(b_l)=b_n/S_n(b)-\sum_mZ_m(x^l)\hat{W}^l_{m,n}\] where \(S(b)=S_{acc}\) defined in Eq. (\ref{eq:acc_scale}) (Appendix \ref{section:ProofOfLR}). This serves as another example of how our general parameterization analysis directly yields quantization formulae previously described as independent special insights. We thus see our rethinking as a step towards a common language for transparent specification/reporting and deployable optimization of any \textit{(HW-scheme + quantization-method)} combination on a single simulator, indispensable for efforts such as \cite{LiMQBench21}.

\section{Proof of Equation (\ref{eq:LRfromAF})}\label{section:ProofOfLR}

We refer to the setup of Fig. \ref{fig:vector_scales_single_conv}, and section \ref{section:DoFConvAnalysis}.
We first introduce the constraint of all terms in the partial sum having the same scale. This constraint is derived from hardware use of multiply-and-accumulate primitive. This is formalized as a \(m\)-invariant encoding of the addends, by a well-defined \textit{accumulator scale} that can be factored out as:

\[ \sum_{m,..}(\hat{A}^{l-1}_{..,
m} * \hat{W}^l_{..,m,n}) = (S_{acc}^l)^{-1}_n 
\sum_{m,..}(A^{l-1}_{..,
m} * W^l_{..,m,n}) \]
\begin{equation}
(S_{acc}^l)_n:=(S^l_w)_{m,n}(S_a^{l-1})_m
\label{eq:acc_scale}
\end{equation}
This constraint mandates that weight-scale tensor is decomposed as follows, defining accumulator scale vector and left/right weight scale co-vectors:
\begin{equation}
 (S^{l}_{w})_{m,n} := (S^{l-1}_a)_m^{-1}(S_{acc}^l)_n := (S^l_{wL})_m * (S^l_{wR})_n
\end{equation}

Thus, weight scale is an outer-product of two co-vectors ("left" and "right"), indexed by input/output channel respectively, rewriting kernel quantization as:
\begin{equation}
 \hat{W^l}_{m,n} = clip\left(\left\lfloor(S^l_{wL})_m^{-1}W^l_{m,n}(S^l_{wR})_n^{-1}\right\rceil\right)
 \label{eq:kernelquant}
\end{equation}
Next, we introduce the relation of scale change upon multiplicative recode: 
\begin{equation}
    (S^l_a)_n = (S_{acc})_n/\hat{F}^l_n
\end{equation}
The last two equations finally enable the expression of kernel scale co-vectors via activation scale vectors and the HW rescale factor (be it vector or scalar): 
\begin{equation}
    (S^l_{wL})_m = (S^{l-1}_a)_m^{-1} \quad ; \qquad (S^l_{wR})_n = (S^l_a)_n*\hat{F}^l_n
    \label{eq:LRfromAF1}
\end{equation}
Thus, given "channelwise" rescale, there are actually \textbf{two} vector degrees of freedom per layer, which can be parameterized via Eq. (\ref{eq:LRfromAF}) or, on some condition, via the "left/right" scales.

This formulation treats input/output channels symmetrically. Our "right" scale vector corresponds to the standard per-[output-]channel quantization step-size. However, the "left" scale vector, novel to this exposition, is equivalent to cross-layer factorization DoF \cite{MellerEqual19,NagelDFQ19} described as preconditioning. We thus rethink this DoF within quantization, as one of two scale co-vectors, constrained by relation to the preceding layer - to its per-channel rescale factors if available, or, in layerwise-quantization case, its kernel right-scales.

\section{Doubly-channelwise MMSE by alternating projections}\label{section:APQ}

For completeness, we first address the simple \textit{scalar-scale}-MMSE (Minimum Mean Square Error) single-variable optimization problem (Eq. (\ref{eq:MMSE_B})), and reproduce in \textit{Algorithm \ref{alg:PPQ}} the iterative solution we adopt from \cite{LiuAlphablend19}. 
%Note that the \(clip\) operator saturates at \(+-2^{b-1}-1\)
\begin{equation}
  MMSE(W):=\min_{s}\left\Vert W - s*clip\left(\left\lfloor W/s\right\rceil\right)\right\Vert 
 \label{eq:MMSE_B}
\end{equation}

\RestyleAlgo{ruled}

The intuition behind the algorithm is that of minimizing error by linear projection. At convergence, the following relation will hold:
\begin{equation}
    0 = \left<sq-x,\ q\right> = \left<e, q\right>
\end{equation}
Where \(e\) is the error of estimation of \(x\) from \(q\). This converged value is therefore optimal - by orthogonality principle for linear estimators. While convergence analysis is beyond our scope here, in practical application for optimal clipping of DNN weight matrices, we observe robust convergence, often after low single-digit number of iterations.

\vspace{-0.3cm} 
\begin{algorithm}
\caption{PPQ - Progressive Projection Quantization (from \cite{LiuAlphablend19})}
\KwData{Full precision vector $\textbf{x}=\{x_i | i\in[0,N-1]\}$}
\KwResult{$x_{mmse} = \underset{s}{\argmin}\left\Vert x - s*clip\left(\left\lfloor x/s\right\rceil\right)\right\Vert$}
$s \gets \dfrac{max(|x|)}{2^{b-1}-1} $\;
\For{each $r \in [0, iterations)$}{
  $q_i \gets x_i/s$ for each $i\in[0,N-1]$\;
  $s \gets \dfrac{\left<x, q\right>}{\left<q, q\right>} = \sum_i{x_iq_i} / \sum_i{q_i^2} $ \;
}
\label{alg:PPQ}
\end{algorithm}
\vspace{-0.3cm} 

Now, we use the same principle to construct a procedure for the \textit{dual vector scaling} MMSE problem, in which instead of a scalar we seek \textit{left} and \textit{right} vector scales. It was used in main text for the application of locally-optimal doubly-channelwise quantization of kernels:
\begin{equation}
\resizebox{0.9\textwidth}{!}{$%
MMSE_{DiCH}(W) = \underset{S_{wL}, S_{wR}}{\min}\left\Vert W_{m,n} - (S_{wL})_m(S_{wR})_n*clip\left(\left\lfloor\frac{W_{m,n}}{(S_{wL})_m(S_{wR})_n}\right\rceil\right)\right\Vert
$}
\label{eq:DiChMMSE}
\end{equation}
We alternate between a single iteration of finding optimal row scale (looped over rows) and finding optimal columns scale (looped over columns). Note the slight asymmetry where we need to take the one scale vector into account when handling the other, since X is scaled by both. The overall procedure is detailed in Algorithm \ref{alg:APQ}. The solution is non-unique, and up to scalar factor movable between \(S\) and \(T\). The overall algorithm is quite fast, even using 10 overall iterations, it takes around a second for matrices sized ~1M (on a strong server).

\begin{algorithm}\label{alg:APQ}
\caption{Alternating Projection Quantization (APQ)}
\KwData{Full precision matrix $\textbf{x}=\{X_{ij} | i\in[0,N-1], j\in[0,M-1]\}$}
\KwResult{$(S,T)_{mmse} = \underset{S_i,S_j}{\argmin}\left\Vert X_{ij} - S_iT_j*clip\left(\left\lfloor\dfrac{X_{ij}}{S_iT_j}\right\rceil\right)\right\Vert$}
$T_j \gets \underset{i}{\max}\left(|X{ij}|\right) /\left(2^{b-1}-1\right) $\;
$S_i \gets \underset{j}{\max}\left(|X{ij}/T_j|\right) /\left(2^{b-1}-1\right) $\;

\For{each $r \in [0, iterations)$}{
$Q_{ij} \gets X_{ij}/\left(S_iT_j\right)$ for each $i\in[0,N-1],j\in[0,M-1]$\;
$T_j \gets \frac{\left<\Tilde{Q_j},\Tilde{X_j}/S\right>}{\left<\Tilde{Q_j}, \Tilde{Q_j}\right>} = \sum_i{Q_{ij}\dfrac{X_{ij}}{S_{ij}}} / \sum_i{Q_{ij}^2} $ for each $j\in[0,M-1] $\;
  
$Q_{ij} \gets X_{ij}/\left(S_iT_j\right)$ for each $i\in[0,N-1],j\in[0,M-1]$\;

$S_i \gets \frac{\left< \Tilde{Q_i},\Tilde{X_i}/T\right>}{\left<\Tilde{Q_i}, \Tilde{Q_i}\right>} = \sum_j{Q_{ij}\dfrac{X_{ij}}{T_{ij}}} / \sum_j{Q_{ij}^2} $ for each $i\in[0,N-1] $\;
}
\end{algorithm}

%\pagebreak
\section{Optimal adaptation of CLE \cite{NagelDFQ19} to 4b-quantized weights}
\label{section:CLE}
The Cross-Layer inverse-Factorization degree of freedom (DoF) is defined in \cite{MellerEqual19,NagelDFQ19} as a pre-conditioning transform on weights, parameterized by \textit{CL-factors} \({C^l_m}\):
\begin{equation}
    \tilde{W}^l_{mn} := W^l_{mn} / C^l_m\ ;\ \ \tilde{W}^{l-1}_{km} := W^{l-1}_{km} * C^l_m
    \label{eq:cle_basic}
\end{equation}
The full-precision network is invariant to switching from \(W^{l-1},W^l\) to \(\tilde{W}^{l-1},\tilde{W}^l\) given a \textit{homogeneous}\footnote{i.e., satisfying \(f(ax)=af(x)\). Note that it doesn't hold for \textit{relu6} activation used in \textit{MobilenetV2}, but in practice (see main text results) the impact of this is low, probably owing to weight distributions heavily skewed towards zero.} 
activation in layer \(l\). The behavior of the network after a uniform layerwise 8b quantization of weights, with scales \(s_w^l=max(abs(W^l))/127\), may however change markedly. That's since the dynamic ranges of the impacted kernel slices \( \{W^{l-1}_{km} | k=1..K\}, \{W^l_{mn} | n=1..N\}\) change in opposite directions, and so does their respective effective precision of representation. The effect is strongest for very different \textit{relative} (w.r.t layerwise) dynamic ranges of corresponding slices are. If weights for a certain layer \(l\) and channel \(m\) satisfy:
 
 %R(W_{l-1},\ fout=m) := & max(abs(\{W_{km}^{l-1}|k=1..K\}))=max(abs(W^{l-1})) / 32.0 \\
 %R(W_l,\ fin=m) := & max(abs(\{W_{mn}^l|n=1..N\}))=max(abs(W^l)) / 2.0
 
 \begin{equation}
\resizebox{0.8\textwidth}{!}{$%
\begin{split}
 R(W_{l-1},\ f_{out}=m) := & max(abs(W^{l-1}_{..,m}))\ \simeq\ max(abs(W^{l-1}))\ /\ 2^5 \\
 R(W_l,\ f_{in}=m) := & max(abs(W^l_{m,..}))\ \simeq\ max(abs(W^l))\ /\ 2^1 \\
\end{split}
$}
\label{eq:toy_cle}
\end{equation}
Then output (input) slice of \(W^{l-1}\) \((W^{l})\) is effectively represented in 3 (7) bits under layerwise-quantization constraint. Thus, \textit{equalizing} by applying Eq. (\ref{eq:cle_basic}) with a factor of \(C^l_m=4.0\) can bring both to \(1/2^3\) of respective layerwise range, effectively quantized in 5 bits, incurring smaller overall error than when dominated by the crude 3b-equivalent quantization of  \(W^{l-1}\). 
This exemplifies the intuition behind the \textit{geometric mean} optimizer of this DoF proposed in \cite{NagelDFQ19} for improving 8b quantization by cross-layer equalizing pre-conditioning of weights\footnote{Note that for simplicity we left out of the scope the impact on activation quantization as well as subtle effects of factoring the strongest channel defining the layerwise range, both calling for modified, possibly multi-staged optimization procedures \cite{MellerEqual19}}.

In main text we reformulate this DoF as a per-channel activation scale \(S^{l-1}_A\) (and derived \textit{left/right} weight scales), reinterpreting the \textit{CLF-factors} as ratios of these vectors' elements to their uniform (layerwise) initialization \footnote{The \(\alpha\) factors provide for a possible post-CLE \textit{scalar re-calibration} within same step}:
\begin{equation}
    (S^{l-1}_A)_m = \alpha_a C^l_m * s^{l-1}_A\ ;\ \ (S^{l-1}_{WR})_m = \alpha_w C^l_m * s^{l-1}_W\ ;\ \ (S^{l}_{WL})_m = 1/C^l_m
    \label{eq:cle_factors_to_scales}
\end{equation}
This fuses the pre-conditioning step(s) into quantization, while having the exact same effect on the eventual quantized network. So, choosing optimal \(C^l_m\) or optimal \((S^{l-1}_A)_m\) is one and the same, but the latter approach lends itself better to joint all-DoF end-to-end training, as shown in main text. 

Now we note that since beginning of section we assumed there's no clipping as a part of the optimal solution. This usually holds for 8-bit quantization where rounding errors are relatively small and even weak clipping losses quickly overwhelm the precision gains from decreasing range. However, for 4-bit it's markedly not the case, and clipping is routinely used, the typical optimal range often in the ballpark of 1/4 of naive \textit{max(abs(W))}\footnote{This 1/4-ratio rule-of-thumb is roughly interpretable as dividing evenly the log-scale \textit{4b gap} of 4b- vs. 8b-quantization: 2b of clipping and 2b of precision loss..}. Percentile-based or even simple 1/4 ratio clipping can give acceptable performance, but principled empirical \textsc{mmse} is preferred. For a typical 'naive' (in the sense of ignoring the CLE DoF) quantization, the \textsc{mmse} is applied to whole tensor W or slice \(\{W_{km}^{l-1}|k=1..K\}\) ) for layerwise or channelwise quantization, respectively, as a replacement of the 8b-optimal \textit{max(abs(.))} on respective tensor/slice.

Crucially, when we add the CLE DoF to the layerwise case\footnote{Note that CLE DoF added to channelwise case yields the decoupled-layers \textit{doubly-channelwise quantization}, see discussion in Section 3.1 and Appendix \ref{section:APQ}. }, 
the \textit{Equalization} and \textit{Clipping} optimization problems become fully coupled, as nontrivial (\(\neq1\)) \textit{CLF-factors} move parts of weights distribution across the clipping threshold. As argued in main text, the conceptually simplest and best performing approach to this joint optimization problem is via a joint training - of vector scales \(S_A\) jointly with rescale factors \(F\) and all other DoF. This results in both optimal \textit{equalization} and optimal \textit{clipping}, as they are simply one and the same!

Here we proceed with the question of what can be done within local heuristic optimization of \textit{CLF-factors} (or, equivalently, \(S_A\)) in the case of per-layer\footnote{The similar question for case of per-channel \(F_n\) leading to doubly-channelwise decoupled-layers clipping optimization is addressed in Appendix \ref{section:APQ}} rescale factors \(F\).
To that end, we formulate the \textit{geometric-mean CLE} \cite{NagelDFQ19} heuristic as a mean between (a.) the CLF-factors optimal for quantization of output-channel \(m\) kernel slice of layer \({l-1}\) and (b.) CLF-factors optimal for quantization of input-channel \(m\) kernel slice of layer \(l\), under layerwise quantization. Either optimum is achieved, in the no-clipping 8-bit case, when the naive-max dynamic range of the slice is brought to meet that of the layer, as exemplified in the toy case study in Eq. (\ref{eq:toy_cle}). As we move to the 4-bit case, we switch from naive-max to \textsc{mmse} and note that CLF-factors bringing \textsc{mmse} kernel-slice scale to \textsc{mmse} full-kernel scale are similarly optimal in the sense of closing the representability gap of layerwise vs. channelwise quantization. Finally, we apply the \textit{geometric-mean} heuristic to the channelwise/layerwise \textsc{mmse} ratios:
\begin{equation}
    2\log\left(C^l_m\right) = \log\left( (\hat{S}^{l-1}_{WR})_m / \hat{s}^{l-1}_{W} \right) + \log\left(\hat{s}^{l}_{W} / (\hat{S}^{l}_{WL})_m\right)
\label{eq:CLE4b}
\end{equation}
Here the hat in \(\hat{x}\) stands for \textit{locally optimal \(x\)}, in the \textsc{mmse} sense (see Eqs.\ref{eq:MMSE}):
\begin{equation}
\resizebox{0.5\textwidth}{!}{$%
\begin{split}
(\hat{S}^{l}_{WL})_m =  \underset{s}{\argmin}\left\Vert W^l_{m,..} - s*clip\left(\left\lfloor\dfrac{W^l_{m,..}}{s}\right\rceil\right)\right\Vert
\\
(\hat{S}^{l-1}_{WR})_m =  \underset{s}{\argmin}\left\Vert W^{l-1}_{..,m} - s*clip\left(\left\lfloor\dfrac{W^{l-1}_{..,m}}{s}\right\rceil\right)\right\Vert
\end{split}
$}    
\end{equation}
We solve these in practice by applying PPQ \cite{LiuAlphablend19} (see Appendix \ref{section:APQ}). Again, this extension actually covers 8b case too, with the \textsc{mmse} typically giving results close to degenerate (no-clipping, naive-max range). Thus, in the 8b case, the first (second) term on RHS of Eq. (\ref{eq:CLE4b}) will be always negative (positive), while in 4b case either term can have either sign. This question of whether optimal range (equivalently, clipping threshold) for kernel slice is higher/lower than that of the whole kernel depends on their respective distributions' widths.
In case of a heterogeneous quantization, e.g. layer \(l-1\) (\(l\)) in 8b (4b), the optimal usage of this DoF is a skew towards the 4b layer's "needs". Stopping short of principled error minimization\footnote{Because of very few 8b layers in our experiments in this work, and more importantly, focus on end-to-end trained optimization}, we suggest a simple heuristic extension as: 
\begin{equation}
    2\log\left(C^l_m\right) = (1+\beta) \log\left( (\hat{S}^{l-1}_{WR})_m / \hat{s}^{l-1}_{W} \right) + (1-\beta) \log\left(\hat{s}^{l}_{W} / (\hat{S}^{l}_{WL})_m\right)
\end{equation}
With higher weighting for lower-bitwidth layers (that is, \(\beta>0\) if \(l-1\) is 4bW and \(l\) is 8bW, and \(\beta<0\) in the converse case. In our experiments we take \(\beta=\pm0.5\) for 8b/4b layers pair). Similar considerations are applied to special layers, e.g. in case layer \(l\) is an \textit{elementwise-add} (ew) taken in this work as full-precision (as we generally stop short of simulating platform-specific lossy-elements), we use \(\beta=1\) effectively using the CLE DoF to the full benefit of layer \(l-1\), assuming the inputs-rescale factors in ew-add implementation playing the role of \(W^l\) are per-channel and have a high enough precision (e.g. 8b or more) to not incur significant loss. Additional caveat of Eq. (\ref{eq:CLE4b}) is a possible fan-out of the \textit{producer} layer \(l-1\) to a few \textit{consumer} layers \(\{l_i|i=1..j\}\); in that case we replace the 2nd term of Eq. (\ref{eq:CLE4b}) by a weighted mean of multiple terms accordingly. 

Importantly, we apply the same constraints/freedoms to our main approach of end-to-end jointly trained CLF-factors (or rather, equivalently, activation vector scales \(S_A\)): 
\begin{enumerate}
    \item In case of elementwise-add \textit{consumer} layer \(l\), we take its inputs-rescale factors playing the role of 'weights' as lossless, thus encouraging (though not directly instructing) the training to use the CLE DoF to full benefit of the \textit{producer} layer's \(l-1\) kernel quantization.
    \item In case of fan-out to multiple \textit{consumer} layers \(\{l_i\}\), we do enforce the constraint of all \(C^{l_i}\) vectors being the same, as they share the same \textit{producer} layer referred to as \(l-1\) in above discussion. In our simulation setup based on vector-scale reformulation this is implemented by simply using the same \(S_A^{l-1}\) for all consumers within the \textit{offline subgraph}.
\end{enumerate}

\begin{figure*}
\vspace{-0.3cm}  
\includegraphics[scale=0.3]{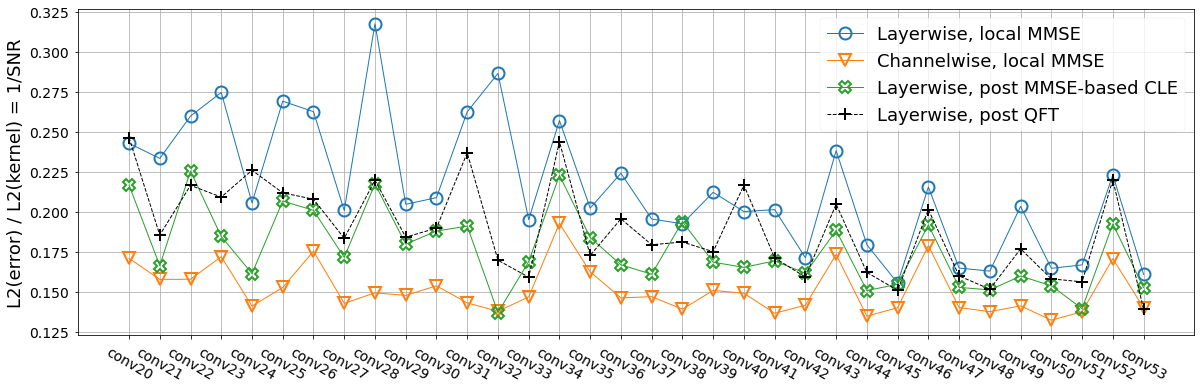} 
\caption{Quantization error for kernels, for different scale-optimization procedures. We see that both CLE and QFT partially close the gap from layerwise to channelwise. The similarity of their effects varies over layers. The kernel error draws a very coarse-grained picture in 2 related respects: (A.) Aggregates un-normalized errors in channels of varying impact or simply magnitude (B.) Only a weak proxy for bottom line network accuracy. In what follows, we drill down into (A) mainly, looking into clipping/rounding errors with a channel resolution.}
\label{fig:quant_error_lwchwcleqft}
%\vspace{-0.2cm}  
\end{figure*}

\begin{figure*}
\vspace{-0.3cm}  
\includegraphics[scale=0.3]{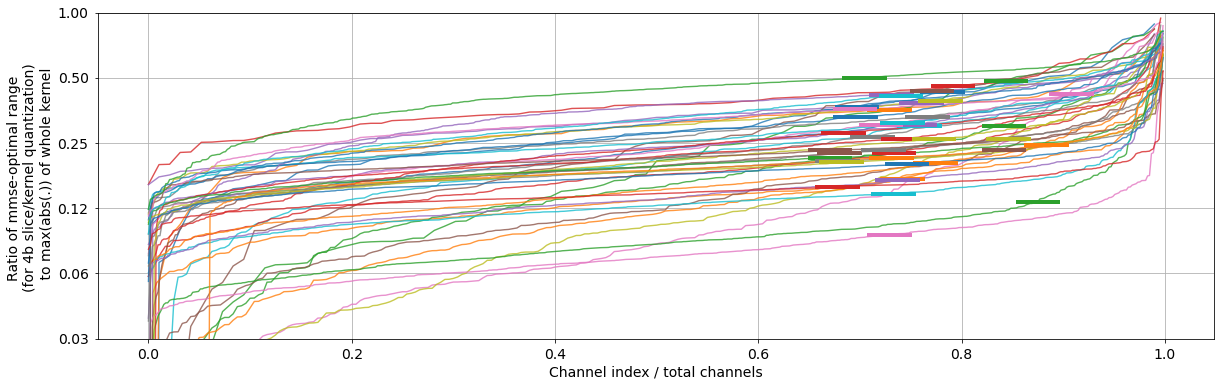} 
\caption{Each color is a layer, each point is a channel, its y-value representing the mmse-optimal range of respective kernel slice under 4b quantization, normalized by naive \(max(abs(.))\) for whole kernel. The bold short horizontal line represents in its y-value the mmse-optimal range of the whole kernel, arbitrarily positioned on x axis to meet the respective per-channel curve. Apparently, very few kernel-slices call for unclipped representation, while for most slices and kernels mmse-optimal 4b quantization is achieved at x2-x8 clipping with respect to }
%\vspace{-0.2cm}  
\end{figure*}

\begin{figure*}
%\vspace{-0.2cm}  
\includegraphics[scale=0.3]{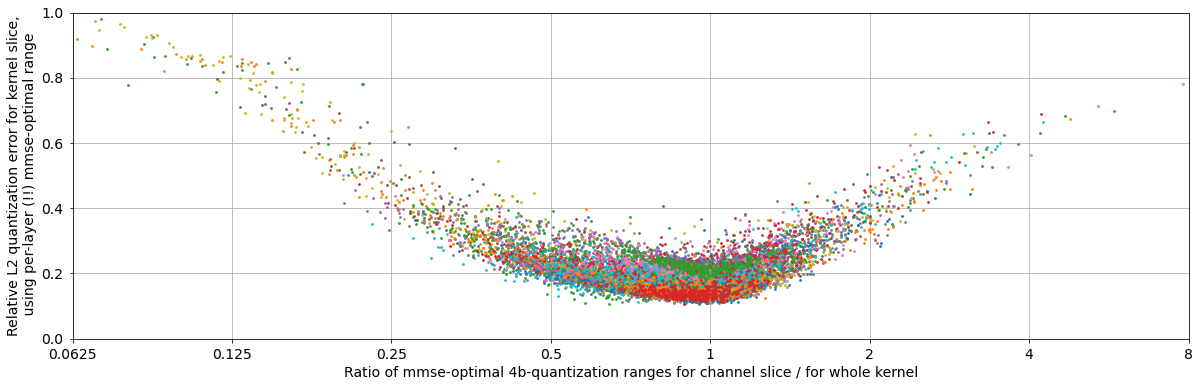} 
\caption{Each color is a layer, each point is a channel, representing the quantization error incurred when representing in 4b this specific kernel slice. Error increases with the deviation, to either side, of kernel-slice optimal range (and underlying that, its distribution) from that of whole kernel. The channels to the right-hand (left-hand) side of x=1 suffer from excessive (inadequate) clipping, respectively, when quantized with uniformly layerwise scale.}
\label{fig:bathtub_layerwise}
%\vspace{-0.3cm}  
\end{figure*}

\begin{figure*}
%\vspace{-0.3cm}  
\includegraphics[scale=0.3]{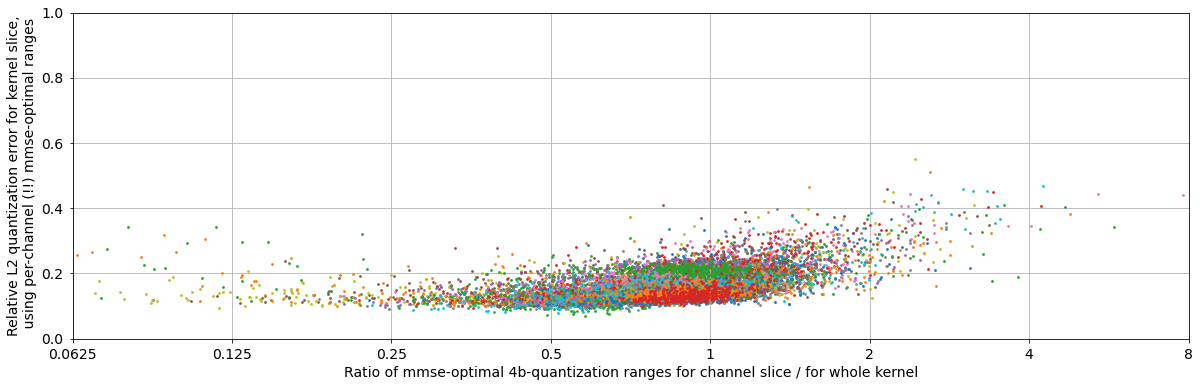} 
\caption{Same as Fig. \ref{fig:bathtub_layerwise}, with channelwise quantization. As expected, lower errors can be achieved for slices deviating in their optimal-mmse range from that of whole kernel.}
%\vspace{-0.3cm}  
\end{figure*}

\begin{figure*}
%\vspace{-0.3cm}  
\includegraphics[scale=0.3]{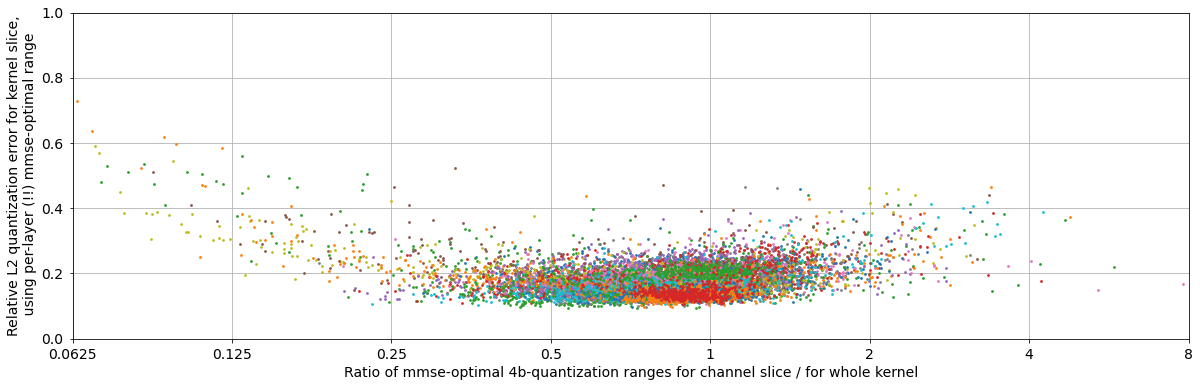} 
\caption{Same as Fig. \ref{fig:bathtub_layerwise}, after 4b-adapted CLE as described in text. Error reduction is smaller but roughly comparable to that of channelwise. Keep in mind that CLE symmetrically accounts for the input channels of next layer.}
%\vspace{-0.5cm}  
\end{figure*}

\begin{figure*}
%\vspace{-0.3cm}  
\includegraphics[scale=0.3]{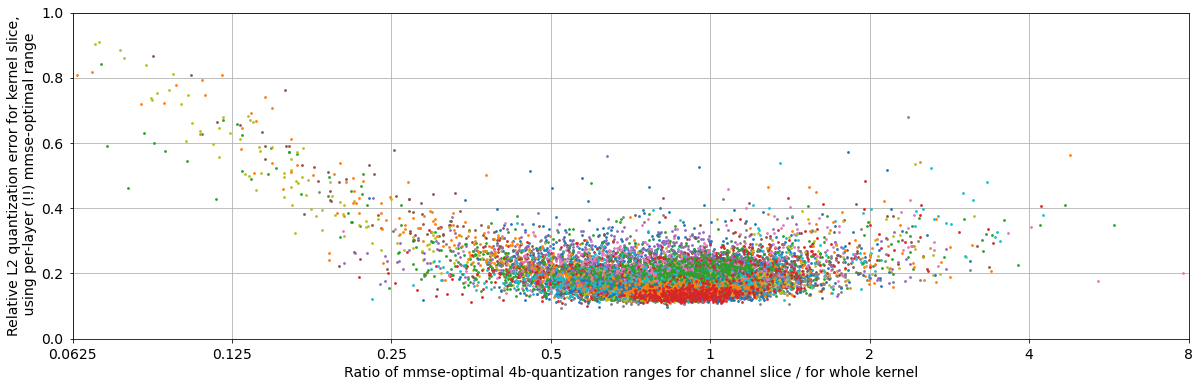} 
\caption{Same as Fig. \ref{fig:bathtub_layerwise}, after QFT. Error reduction is slightly smaller than with CLE but still roughly comparable to that of channelwise, and here achieved without explicit heuristics and even without explicit optimization of kernel quantization error. Rather, data-driven minimization of the network-output compound quantization error is used in QFT. The mechanism for downstream error reduction through CLE DoF tuning is still through less info loss in clipping, but e.g. the input/output-channel balance could be tilted away from local optimum in favor of downstream objective.}
%\vspace{-0.5cm}  
\end{figure*}

\subsubsection{Channelwise vs. CLE DoF at a glance, analytics for 4b case}
Summarizing the overall journey of the CLE DoF from \cite{MellerEqual19,NagelDFQ19} and through this work:
\begin{enumerate}
    \item Any uniform quantization scheme suffers from discrepancy between the optimal representation ranges called for by different sub-tensors.
    \item Channelwise vs. Layerwise 4b-quantization of conv kernels is an example, with the former alleviating the "desired ranges discrepancy" across output channels' (but not input channels') slices of kernel.
    \item The CLE DoF enables a partial per-slice range flexibility at a zero HW cost but inversely coupled within cross-layer pairs of output/input channel slices.
    \item The DoF can be either optimized towards local error minimization with a closed-form heuristic, or better, end-to-end trained, approaching the accuracy gain of channelwise quantization.
\end{enumerate}
We illustrate the above with analyses of the kernels of \textit{regnetx600mf} net, in Fig. \ref{fig:quant_error_lwchwcleqft} and subsequent figures.

\pagebreak
\section{Accuracy without QFT}

\begin{table}[]
\caption{Ablation: using heuristic local scale optimizations and stopping short of running QFT itself. \textsc{mmse} stands for layerwise and doubly-channelwise mmse-optimal ranges, according to the setting. \textsc{cle} stands for 4b-adapted cross-layer equalization as described in Appendix \ref{section:CLE}. \textsc{bc*} stands for empirical bias correction \cite{FinkelIBC19}, optionally applied, with best results chosen out of w./w.o. BC. }
\resizebox{\textwidth}{!}{%

\begin{tabular}{@{}lllllllll@{}}
\toprule
               &            &  & \multicolumn{6}{l}{ImageNet-1K accuracy (degradation)}                    \\ \midrule
Methods        & bits (W/A) &  & ResNet18 & MobileNetV2 & RegNet0.6G & MnasNet2    & ResNet50 & RegNet3.2G \\ \midrule
Full Precision & 32/32      &  & 71.2     & 72.8        & 73.8       & 76.7        & 76.8     &            \\ \midrule
mmse+bc        & 4/8, lw    &  & 30 (-41) & 0.1 (-72.6) & 33.8 (-40) & 69.6 (-7)   & 53 (-24) &            \\
mmse+CLE+bc    & 4/8, lw    &  & 47 (-24) & 0.1 (-72.6) & 49.5 (-24) & 72.1 (-4.5) & 59 (-18) &            \\
mmse+CLE+\textbf{QFT} & 4/8, lw   &  & 70.35 (-0.9) & 72.0 (-0.8) & 72.6 (-1.2)  & 76.35 (-0.3) & 76.2 (-0.6)   & 77.7 (-0.8)  \\ \midrule
mmse+bc               & 4/32, chw &  & 57.5 (-14)   & 0.1 (-72.6) & 63 (-10.7)   & 71.2 (-5.4)  & 69.5 (-7.3)   & 70.8 (-7.7)  \\
mmse+\textbf{QFT}     & 4/32, chw &  & 70.8 (-0.45) & 71.9 (-0.9) & 73.0 (-0.85) & 76.2 (-0.45) & 76.45 (-0.35) & 78.15(-0.35)
\end{tabular}
}
\label{table:no-qft}
\end{table}

Last but not least, to fully appreciate the added value of QFT we explore the best accuracy that can be reached without modifying weights, only heuristically optimizing the ranges (and possibly, biases). This can be seen as complementary to Figs. \ref{fig:CLE_vs_QFT},\ref{fig:doubly_QFT}, in that we ablate by abstain from optimizing some of the DoF, here the weights themselves. We use the same settings as in Experiments section of the main text. Contrary to experiments reported there, we stop short of repeated runs as the degradation results, as can be seen in Table \ref{table:no-qft} are two orders of magnitudes above the typical variability of roughly 0.2\%. We can see that such quantization yields a massive to complete (\textit{MobileNetV2}) loss of accuracy, only partially alleviated by CLE and even doubly-channelwise setting. When comparing to QFT results in main text, we see degradation reduction of typically x10-30 thanks to finetuning the weights themselves (jointly with scales and biases). The contribution of weights training is thus, as expected, much more critical that that of scales training, which supplies the final x2 degradation reduction pushing the final result to be below the 1\% mark and on-par or beyond SoTA (Figs. \ref{fig:CLE_vs_QFT},\ref{fig:doubly_QFT}). Crucially, with our QFT this critical finetuning of the weights is very fast and entails neither compute and data-intensive training as in classic QAT nor layer-by-layer operation and specially designed procedure as in \cite{LiBRECQ21,NagelAdaround20}, nor per-network configuration ubiquitously used by either methods' class. 

\end{document}